# Intergration Of Visual Inter-word Constraints And Linguistic Knowledge In Degraded Text Recognition

Tao Hong
Center of Excellence for Document Analysis and Recognition
Department of Computer Science
State University of New York at Buffalo, Buffalo, NY 14260
taohong@cs.buffalo.edu

## Abstract

Degraded text recognition is a difficult task. Given a noisy text image, a word recognizer can be applied to generate several word candidates for each word image. High-level knowledge sources can then be used to select a candidate from the candidate set for each word image. In this paper, we propose that visual inter-word constraints can be used to facilitate candidate selection. Visual inter-word constraints provide a way to link word images inside the text page, and to interpret them systematically.

## Introduction

The objective of visual text recognition is to transform an arbitrary image of text into its symbolic equivalent correctly. Recent technical advances in the area of document recognition have made automatic text recognition a viable alternative to manual key entry. Given a high quality text page, a commercial document recognition system can recognize the words on the page at a high correct rate. However, given a degraded text page, such as a multiple-generation photocopy or facsimile, performance usually drops abruptly([1]).

Given a degraded text image, word images can be extracted after layout analysis. A word image from a degraded text page may have touching characters, broken characters, distorted or blurred characters, which may make the word image difficult to recognize accurately. After character recognition and correction based on dictionary look-up, a word recognizer will provide one or more word candidates for each word image. Figure 1 lists the word candidate sets for the sentence, "*Please fill in the application form.*" Each word candidate has a confidence score, but the score may not be reliable because of noise in the image. The correct word candidate is usually in the candidate set, but may not be the candidate with the highest confidence score. Instead of simply picking up the word candidate with the highest recognition score, which may make the correct rate quite low, we need to find a method which can select a candidate for each word image so that the correct rate can be as high as possible.

Contextual information and high-level knowledge can be used to select a decision word for each word image

```
1         2      3      4      5              6       7
Please    fin    in     tire   application    farm    !
0.90      0.33   0.30   0.80   0.90           0.35
Fleece    fill   In     toe    applicators    form
0.05      0.30   0.28   0.10   0.05           0.30
Pierce    flu    lo     lire   acquisition    forth
0.02      0.21   0.25   0.05   0.03           0.20
Fierce    flit   ill    the    duplication    foam
0.02      0.10   0.13   0.03   0.01           0.11
Pieces    till   Io     Ike    implication    force
0.01      0.06   0.04   0.02   0.01           0.04
```

Figure 1: Candidate Sets for the Sentence: "*Please fill in the application form!*"

in its context. Currently, there are two approaches, the statistical approach and the structural approach, towards the problem of candidate selection. In the statistical approach, language models, such as a *Hidden Markov Model* and word collocation can be utilized for candidate selection ([2, 4, 5]). In the structural approach, lattice parsing techniques have been developed for candidate selection([3, 7]).

The contextual constraints considered in a statistical language model, such as word collocation, are local constraints. For a word image, a candidate will be selected according to the candidate information from its neighboring word images in a fixed window size. The window size is usually set as 1 or 2. In the lattice parsing method, a grammar is used to select a candidate for each word image inside a sentence so that the sequence of those selected candidates form a grammatical and meaningful sentence. For example, consider the sentence "*Please fill in the application form*". We assume all words except the word "*form*" have been recognized correctly and the candidate set for the word "*form*" is { *farm, form, forth, foam, forth* } (see the second sentence in Figure 2). The candidate "*form*" can be selected easily because word collocation between the words "*application*" and "*form*" is strong and the sentence is grammatical by choosing the candidate "*form.*"

The contextual information inside a small window or inside a sentence sometimes may not be enough to select a candidate correctly. For example, consider the sentence "*This form is almost the same as that one*"(see the first sentence in Figure 2). The word image 16 has five candidates: { *farm, form, forth, foam, forth* }. After lattice parsing, the candidate "*forth*" will be removed because it does not fit the context. But it is difficult to select a candidate from "*farm*", "*form*" "*foam*" and "*force*" because each of them makes the sentence grammatical and meaningful. In such a case, more contextual constraints are needed to distinguish the remaining candidates and to select one.

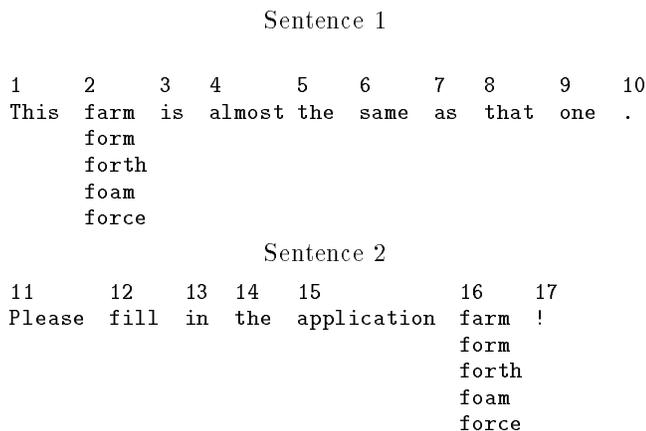

Figure 2: Word candidates of two example sentences(word images 2 and 16 are similar)

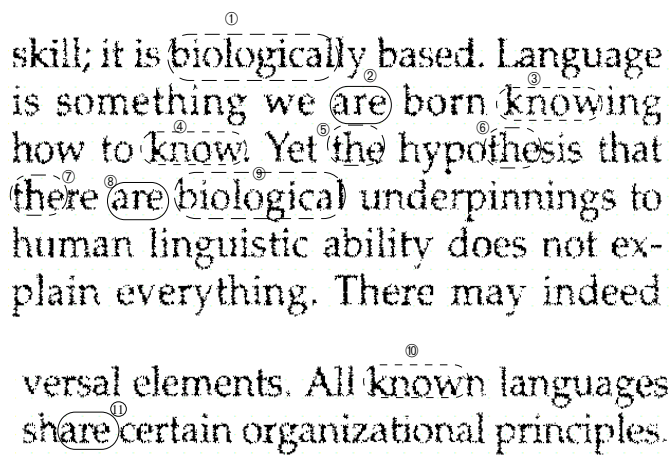

Figure 4: Part of text page with three sentences

Let's further assume that the sentences in Figure 2 are from the same text. By image matching, we know word images 2 and 16 are visually similar. If two word images are almost the same, they must be the same word. Therefore, for the word image *2* and the word image *16*, same candidates must be selected. After the word image *16* chooses "*form*", the image *2* will also choose "*form*" as its identity.

|  | Possible Relations between $W_1$ and $W_2$ | |
|---|---|---|
| type | at symbolic level | at image level |
| 1 | $W_1 = W_2$ | $W_1 \approx W_2$ |
| 2 | $W_2 = X \bullet W_1 \bullet Y$ | $W_1 \approx subimage\_of(W_2)$ |
| 3 | $prefix\_of(W_1) =$ $prefix\_of(W_2)$ | $left\_part\_of(W_1) \approx$ $left\_part\_of(W_2)$ |
| 4 | $suffix\_of(W_1) =$ $suffix\_of(W_2)$ | $right\_part\_of(W_1) \approx$ $right\_part\_of(W_2)$ |
| 5 | $suffix\_of(W_1) =$ $prefix\_of(W_2)$ | $right\_part\_of(W_1) \approx$ $left\_part\_of(W_2)$ |

Note 1: "$\approx$" means approximately image matching;
Note 2: "$\bullet$" means concatenation.

Figure 3: Possible Inter-word Relations

## Visual Inter-Word Relations

A visual inter-word relation can be defined on two word images if they share the same pattern at the image level.

There are 5 types of visual inter-word relations listed in the right part of Figure 3. Figure 4 is a part of a scanned text image in which a small number of word relations are circled to demonstrate the abundance of inter-word relations defined above even in such a small fragment of a real text page. Word images 2 and 8 are almost the same. Word image 9 can match the left part of the word image 1 quite well. Word image 5 can match a part of the image 6, and so on.

Visual inter-word relations can be computed by applying simple image matching techniques. They can be defined in clean text images, as well as in highly degraded text images, because the word images, due to their relatively large size, are quite tolerant to noise ([6]).

Visual inter-word relations can be used as constraints in the process of word image interpretation, especially for candidate selection. It is not surprising that word relations at the image level are highly consistent with word relations at the symbolic level(see Figure 3). *If two words hold a relation at the symbolic level and they are written in the same font and size, their word images should keep the same relation at the image level.* And also, *if two word images hold a relation at the image level, the truth values of the word images should have the same relation at the symbolic level.* In Figure 4, word images 2 and 8 must be recognized as the same word because they can match each other; the identity of word image 5 must be a sub-string of the identity of word image 6 because word image 5 can match with a part of word image 6; and so on.

Visual inter-word constraints provide us a way to link word images inside a text page, and to interpret them systematically. Integrating visual inter-word constraints with a statistical language model and lattice parser improves the performance of candidate selection, as shown above.

## Current Status of Work

A word-collocation-based relaxation algorithm and a probabilistic lattice chart parser have been designed for word candidate selection in degraded text recognition([3, 4]). The relaxation algorithm runs iteratively. In each iteration, the confidence score of each candidate is upgraded based on its current confidence and its word collocation scores with currently most preferred candidates of its neighboring word images. Relaxation ends when all candidates reach their stable state. For each word image, those candidates with low confidence score will be removed from candidate sets. Then, the probabilistic lattice chart parser will be applied to the reduced candidate sets to select the candidates which appear in the most preferred parse trees built by the parser. There can be different strategies to use visual inter-word constraints inside the relaxation algorithm and the lattice parser. One of the strategies we are exploiting is to re-evaluate the top1 candidates of the related word images after each iteration of relaxation or after lattice parsing. If they hold same relation at symbolic level, the confidence scores of the candidates will be increased. Otherwise, the images with low confidence score will follow the decision of the images with high confidence score.

Five articles from the Brown Corpus were chosen randomly as testing samples. They are *A06*, *G02*, *J42*, *N01* and *R07*, each with about 2,000 words. Given a word image, our word recognizer can generate its top10 candidates from a dictionary with 70,000 different entries. In our preliminary experiment, we exploit only the *type-1* relation listed in Figure 3. After clustering word images by image matching, similar images will be in same cluster. Any two images from same cluster hold the *type-1* relation. Word collocation data were trained from the Penn Treebank and the Brown Corpus except five testing samples. Figure 5 shows results of candidate selection with and without using visual inter-word constraints. The top1 correct rate of candidate lists generated by a word recognizer is as low as 57.10%, Without using visual inter-word constraints, the correct rate of candidate selection by relaxation and lattice parsing is 83.19%. After using visual inter-word constraints, the correct rate becomes 88.22%.

| Article | Number Of Words | Word Recognition Result | Candidate Selection Using No Constraints | Candidate Selection Using Constraints |
|---|---|---|---|---|
| A06 | 2213 | 53.82% | 83.15% | 88.52% |
| G02 | 2267 | 67.71% | 83.81% | 87.87% |
| J42 | 2269 | 54.52% | 83.65% | 89.51% |
| N01 | 2313 | 57.33% | 82.71% | 87.12% |
| R07 | 2340 | 52.22% | 82.69% | 88.16% |
| Total | 11402 | 57.10% | 83.19% | 88.22% |

Figure 5: Comparison Of Candidate Selection Results

## Conclusions and Future Directions

Integration of natural language processing and image processing is a new area of interest in document analysis. Word candidate selection is a problem we are faced with in degraded text recognition, as well as in handwriting recognition. Statistical language models and lattice parsers have been designed for the problem. Visual inter-word constraints in a text page can be used with linguistic knowledge sources to facilitate candidate selection. Preliminary experimental results show the performance of candidate selection is improved significantly although only one type of inter-word relations was used. To fully integrate visual inter-word constraints and linguistic knowledge sources in the relaxation algorithm and the lattice parser is the next step to try.

## Acknowledgments

I would like to thank Jonathan J. Hull for his support and his helpful comments on drafts of this paper.

## References

[1] Henry S. Baird, "Document Image Defect Models and Their Uses," in *Proceedings of the Second International Conference on Document Analysis and Recognition ICDAR-93*, Tsukuba, Japan, October 20-22, 1993, pp. 62-67.

[2] Kenneth Ward Church and Patrick Hanks, "Word Association Norms, Mutual Information, and Lexicography," *Computational Linguistics*, Vol. 16, No. 1, pp. 22-29, 1990.

[3] Tao Hong and Jonathan J. Hull, "Text Recognition Enhancement with a Probabilistic Lattice Chart Parser," in *Proceedings of the Second International Conference on Document Analysis and Recognition ICDAR-93*, Tsukuba, Japan, October 20-22, 1993.

[4] Tao Hong and Jonathan J. Hull, "Degraded Text Recognition Using Word Collocation," in *Proceedings of IS&T/SPIE Symposium on Document Recognition*, San Jose, CA, February 6-10, 1994.

[5] Jonathan J. Hull, "A Hidden Markov Model for Language Syntax in Text Recognition," in *Proceedings of 11th IAPR International Conference on Pattern Recognition*, The Hague, The Netherlands, pp.124-127, 1992.

[6] Siamak Khoubyari and Jonathan J. Hull, "Keyword Location in Noisy Document Image," in *Proceedings of the Second Annual Symposium on Document Analysis and Information Retrieval*, Las Vegas, Nevada, pp. 217-231, April 26-28, 1993.

[7] Masaru Tomita, "An Efficient Word Lattice Parsing Algorithm for Continuous Speech Recognition," in *Proceedings of the International Conference on Acoustic, Speech and Signal Processing*, 1986.